\documentclass[11pt,a4paper]{article}
\usepackage{times,latexsym}
\usepackage{url}
\usepackage[T1]{fontenc}
\usepackage{enumitem} 
\usepackage{graphicx}
\usepackage{microtype}
\usepackage{amsmath}
\usepackage{amsfonts}
\usepackage{amssymb}
\usepackage{subcaption}  
\usepackage{tikz}
\usetikzlibrary{arrows.meta}
\definecolor{pushcol}{HTML}{C25B43}   %
\definecolor{holdcol}{HTML}{4F8C82}   %
\definecolor{dotedge}{HTML}{8B8580}   %
\colorlet{bandcol}{pushcol!9}
\usepackage{makecell}

\usepackage[acceptedWithA]{tacl2021v1}

\usepackage{xspace,mfirstuc,tabulary}

\newif\iftaclinstructions
\taclinstructionsfalse %
\iftaclinstructions
\renewcommand{\confidential}{}
\renewcommand{\anonsubtext}{(No author info supplied here, for consistency with
TACL-submission anonymization requirements)}
\newcommand{\instr}
\fi

\iftaclpubformat %

\else

\fi

\title{Decomposing Factual Sycophancy in Language Models: How Size and Instruction Tuning Shape Robustness}

\author{
  Victor De Marez%
  \and
  Luna De Bruyne
  \and
  Walter Daelemans
  \\
  \ \\
  Centre for Computational Linguistics, Psycholinguistics and Sociolinguistics
  \\
  University of Antwerp
  \\
  Antwerp, Belgium
  \\
  \texttt{firstname.lastname@uantwerpen.be}
  \\
}

\date{}

\begin{document}
\maketitle
\begin{abstract}
Factual sycophancy occurs when a language model abandons a correct, verifiable answer under social pressure. Because a flip occurs only when pressure toward a false answer exceeds the model's neutral preference for the truth, flip rates conflate two mechanisms: the strength of that baseline preference (\textit{truth margin}), and how far pressure shifts it (\textit{manipulation sensitivity}). We decompose factual sycophancy into these channels and use them to separate the effects of size and instruction tuning across 56 open-weight models spanning 0.3B-32B parameters and 13 manipulation types. We find that vulnerability is governed mainly by size, but instruction tuning changes how size acts: small instruction-tuned models can become less robust, whereas large instruction-tuned models usually become more robust. Instruction tuning primarily increases truth margin, but its behavioral effect depends on manipulation type. Scaling also changes the two channels differently: base models gain margin but become mildly more manipulation-sensitive, whereas instruction-tuned models gain margin faster and become less sensitive. Factual sycophancy is therefore not a single scalar property. Evaluations should report channel-specific, manipulation-specific, and size-conditioned robustness rather than flip rates alone.

\end{abstract}

Large language models (LLMs) are increasingly optimized to be \textit{aligned}, that is, to be helpful, honest, and harmless \cite{askell2021general}. A known side effect of this goal is \textbf{sycophancy}, where LLM outputs look preferable to users or follow the user's view, rather than actually improving the responses \cite{perez2023discovering}. We differentiate between two forms \cite{panickssery2023reducing}. \textit{Opinion sycophancy} concerns subjective preferences, where various opinions are acceptable, making assessment difficult. Because users may legitimately want responses tailored to their tastes, values, or viewpoints, agreement on subjective matters is not always undesirable, but excessive or unwarranted validation can still be harmful \cite{doi:10.1126/science.aec8352}. \textit{Factual sycophancy}, by contrast, concerns verifiable claims: deference to a user's false belief is a reliability failure. The safety relevance is direct in domains where misinformation is harmful, such as finance and medicine. It also aligns with governmental trustworthiness frameworks that prioritize reliability, accuracy, and robustness \cite{Tabassi2023-rg, constantino2024article}.

Factual sycophancy can be affected by multiple parts of the model-development pipeline, from pretraining data to supervised fine-tuning and preference optimization \cite{malmqvist2025sycophancy, sharma2023towards}. Existing work has implicated model confidence, size, post-training, and manipulation type in factual sycophancy \cite{wei2023simple, ren2025mask, ranaldi2023large, zhang2025sycophancy, ccelebi2025parrot}. However, these factors are usually studied in isolation. This makes it difficult to draw actionable conclusions from existing results: without crossing them in the same experimental setting, we cannot determine whether observed robustness differences reflect size, instruction-tuning status, architecture, attack type, or their interactions.

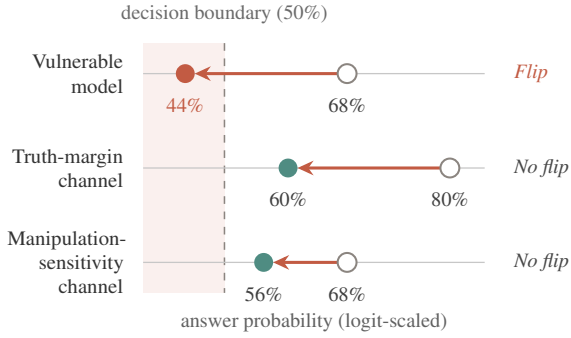
\begin{figure}[!tbp]
\centering
\resizebox{\columnwidth}{!}{%
\begin{tikzpicture}[
    x=2.24cm, y=1cm,
    rowhdr/.style = {font=\scriptsize, color=black!80, anchor=east,
                     align=right, text width=1.75cm, inner sep=0pt},
    pct/.style     = {font=\scriptsize, color=black!75, anchor=north},
    pctflip/.style = {font=\scriptsize, color=pushcol,  anchor=north},
    verdict/.style = {font=\scriptsize\itshape,          anchor=west},
    bdry/.style    = {font=\scriptsize, color=black!58, anchor=south},
    axislbl/.style = {font=\scriptsize, color=black!62, anchor=center},
    axisline/.style= {color=black!22, line width=0.5pt},
    push/.style    = {pushcol, line width=1.2pt,
                      -{Stealth[length=2.2mm,width=2.0mm]}},
    hollow/.style  = {circle, draw=dotedge, line width=0.8pt, fill=white,
                      inner sep=0pt, minimum size=2.6mm},
    flipdot/.style = {circle, fill=pushcol, inner sep=0pt, minimum size=2.6mm},
    holddot/.style = {circle, fill=holdcol, inner sep=0pt, minimum size=2.6mm},
  ]
 
  \fill[bandcol] (-0.5,0.42) rectangle (0,-3.02);
  \draw[dotedge, dashed, line width=0.6pt] (0,0.42) -- (0,-3.02);
  \node[bdry] at (0,0.56) {decision boundary (50\%)};
 
  \draw[axisline] (-0.5,0) -- (1.6,0);
  \node[rowhdr] at (-0.62,0) {Vulnerable model};
  \draw[push]   (0.754,0) -- (-0.183,0);
  \node[hollow]  at (0.754,0)  {};
  \node[flipdot] at (-0.241,0) {};
  \node[pct]     at (0.754,-0.2)  {68\%};
  \node[pctflip] at (-0.241,-0.2) {44\%};
  \node[verdict, color=pushcol] at (1.72,0) {Flip};
 
  \draw[axisline] (-0.5,-1.3) -- (1.6,-1.3);
  \node[rowhdr] at (-0.62,-1.3) {Truth-margin \\ channel};
  \draw[push]   (1.386,-1.3) -- (0.449,-1.3);
  \node[hollow]  at (1.386,-1.3) {};
  \node[holddot] at (0.391,-1.3) {};
  \node[pct] at (1.386,-1.5) {80\%};
  \node[pct] at (0.391,-1.5) {60\%};
  \node[verdict, color=black!72] at (1.72,-1.3) {No flip};
 
  \draw[axisline] (-0.5,-2.6) -- (1.6,-2.6);
  \node[rowhdr] at (-0.62,-2.6) {Manipulation-sensitivity channel};
  \draw[push]   (0.754,-2.6) -- (0.299,-2.6);
  \node[hollow]  at (0.754,-2.6) {};
  \node[holddot] at (0.241,-2.6) {};
  \node[pct] at (0.754,-2.8) {68\%};
  \node[pct] at (0.241,-2.8) {56\%};
  \node[verdict, color=black!72] at (1.72,-2.6) {No flip};
 
  \node[axislbl] at (0.55,-3.42) {answer probability (logit-scaled)};
 
\end{tikzpicture}%
    }
\caption{\textbf{Factual sycophancy is a threshold event, decomposable into two channels.} A \emph{flip} occurs when a sycophantic \emph{push} (orange) carries a model's confidence in the true answer past the $50\%$ decision boundary, and is averted either by a higher baseline margin (\emph{truth-margin channel}, middle) or by a smaller push (\emph{manipulation-sensitivity channel}, bottom). Percentages are illustrative.}
\label{fig:concept}
\end{figure}

Further, sycophancy results are usually reported as outcome metrics, such as accuracies or flip rates. A flip is a threshold event: it occurs when the manipulation-induced shift toward a false answer exceeds the model's baseline margin for the truth. Robustness therefore decomposes into two channels: a \textbf{truth-margin channel} (how strongly the model favors the truth at baseline, which sets how high the threshold sits) and a \textbf{manipulation-sensitivity channel} (how far a given push shifts that preference), as illustrated in Figure \ref{fig:concept}. Knowing which channel an intervention should target requires measuring the channels directly, not just how model attributes relate to flip rates.

We address these gaps through three questions.
\begin{enumerate}[label=\textbf{(RQ\arabic*)}]
\item Which \textbf{manipulation types} most effectively induce factual sycophancy, and how much does effectiveness vary across types?
\item How do \textbf{model size, instruction-tuning status, architecture family}, and their interactions predict worst-case sycophancy vulnerability?
\item Through which channels, \textbf{truth margin or manipulation sensitivity}, do size and instruction-tuning status affect sycophancy, and do those channels account for their interaction?
\end{enumerate}

\noindent This paper makes four contributions:
\begin{enumerate}[leftmargin=1.4em, itemsep=2pt, topsep=3pt, parsep=0pt]
  \item \textbf{A controlled factorial evaluation.}
        We cross model size, instruction-tuning status, architecture family, and 13 manipulation types on the same 56 open-weight checkpoints (0.3B-32B, six families), filtered in two stages for knowledge and competence, allowing to separate the effect of each factor (Section \ref{sec:design})\footnote{The data and evaluations are available at \url{https://github.com/Victordmz/decomposing-factual-sycophancy}}.
  \item \textbf{A unified manipulation ranking.}
        We give, to our knowledge, the first head-to-head comparison of a diverse set of sycophancy manipulations under a single metric, and find that effectiveness is tiered across types (Section \ref{sec:rq1}).
  \item \textbf{A channel decomposition of factual sycophancy.}
        We reframe a flip as a threshold event that occurs only when the manipulation-induced push exceeds the model's baseline truth margin, and give an additive, logit-scale decomposition of robustness into a \emph{truth-margin} channel and a \emph{manipulation-sensitivity} channel, separating mechanisms that flip rates alone conflate (Section \ref{sec:metrics}).
 \item \textbf{An account of how size and instruction tuning shape robustness.} Applying the decomposition, we show that worst-case vulnerability is structured mainly along the size axis, including  its interaction with instruction tuning, whose effect on robustness reverses with size; that instruction tuning improves robustness chiefly by enlarging the truth margin; and that the two channels scale differently across base and instruction-tuned checkpoints, which accounts for most of that interaction (Sections \ref{sec:rq2}-\ref{sec:rq3}).
\end{enumerate}

\section{Related work}
We organize prior work by which component of the flip condition each study illuminates: the model's baseline truth margin (Section~\ref{sec:confidencebuffer}), the manipulation-induced shift (Section~\ref{sec:strategies}), and model attributes that affect both (Section~\ref{sec:scale_alignment}). Some studies bear on more than one component.

\subsection{The confidence buffer}\label{sec:confidencebuffer}
One line of work links sycophantic compliance to baseline certainty. \citet{anagnostidis2024how} find models more resistant to incorrect suggestions on items with high baseline accuracy. \citet{wei2023simple} treat prior factual correctness as a prerequisite for measuring factual sycophancy. The relationship is not deterministic: \citet{sharma2023towards} show GPT-4 still switches answers at 98.9\% stated confidence.

Two recent benchmarks separate behavioral compliance from underlying belief. PARROT \citep{ccelebi2025parrot} records, per item under a false expert claim, both the answer flip and the confidence shift on the true and asserted-false answers, and assigns each outcome to one of eight behavioral categories. MASK \citep{ren2025mask} separates accuracy (belief vs.\ ground truth) from honesty (statement vs.\ belief under pressure) across thirty frontier models. We instead formalize the separation as a flip condition and apply it within a controlled factorial design.

\subsection{Manipulation types and their differential effects}\label{sec:strategies}
Prior work uses four broad families of social pressure: authority and expertise framings \citep{ccelebi2025parrot, anagnostidis2024how, wang2025truth}, user-belief and confidence injection \citep{sharma2023towards, wei2023simple, ranaldi2023large}, escalating or multi-turn pressure \citep{fanous2025syceval, hong2025measuring}, and deception-incentive framings \citep{ren2025mask}. \citet{perez2023discovering} additionally prepend first-person user biographies. These manipulations differ in form, and likely in push magnitudes, but to our knowledge no prior study compares a diverse set of them on the same models under a single metric.

\subsection{Aggregate effects of size and post-training on factual sycophancy}\label{sec:scale_alignment}
A third line of work probes size and post-training, reporting flip rates and accuracies without decomposing them. Findings on size are mixed. \citet{ccelebi2025parrot} observe size-linked improvement within model families across 22 models under expert-authority manipulation, and \citet{ranaldi2023large} report that smaller or lower-accuracy models more often follow misleading users on QA and other verifiable tasks. By contrast, \citet{anagnostidis2024how} find larger Llama-2 chat models \emph{more} easily influenced by multiple-choice explanations, and \citet{zhang2025sycophancy} report no clear size correlation across families under embedded-stance and multi-turn dialogue. \citet{wei2023simple} report every Flan-PaLM size flipping under combined authority-and-belief pressure despite near-perfect neutral accuracy.

On post-training, \citet{sharma2023towards} find that matching a user's stated views is among the most predictive features of human preference in an RLHF preference dataset, and that optimizing responses against their preference model can increase several forms of sycophancy. \citet{ren2025mask} find scaling improves accuracy but not honesty across thirty frontier models. \citet{hong2025measuring} compare Base and IT variants of three open families and find no clear post-training effect.

These studies vary in experimental setup, so their size and post-training findings cannot be compared directly. Separating the size and instruction-tuning effects, and asking whether they act through the baseline preference for the truth or through sensitivity to pressure, requires a controlled factorial design over many families, paired with a decomposition of the flip itself.

\section{Methodology}
\subsection{Experimental design and dataset}\label{sec:design}
We evaluate 56 language models on multiple-choice question-answering (MCQA), spanning six families (OLMo2, Gemma 2, Qwen 2.5, LLaMA 3.2, Qwen 3, Gemma 3) and parameter counts from 0.3B to 32B. Across families, available regimes include pretrained (\textit{Base}), supervised fine-tuned (\textit{SFT}), direct preference optimized (\textit{DPO}), and instruction-tuned (\textit{IT}). Only OLMo2 releases all four at matched size. Our operative contrast is \textbf{instruction-tuning status}, \textit{Base} vs.\ instruction-tuned (\textit{IT}), the only regime distinction available across all six families. \textit{IT} denotes the endpoint of the vendor's post-training pipeline. The paired and channel analyses (Section~\ref{sec:rq3}) use this Base-IT contrast directly, whereas the pooled manipulation ranking (Section~\ref{sec:rq1}) and the model-level regression (Section~\ref{sec:rq2}) additionally include the intermediate OLMo2 SFT and DPO checkpoints as regime levels, which we do not analyze as a separate mechanism. %

Stimuli are drawn from PlausibleQA, which pairs each question with several plausible bait answers \cite{mozafari2025wrong}. For each question $q$ we form five truth-bait pairings $(a, b_1), \ldots, (a, b_5)$ from the gold answer $a$ and five baits. Each pairing is presented as a two-option multiple-choice prompt containing the question, the lettered options, any manipulation text, and a trailing \texttt{Answer: (} for the model to complete. Chat-template models receive a role change before the final line for evaluation in their native elicitation format.%

To ensure we measure the corruption of existing factual knowledge, we apply two filters:
\begin{itemize}
    \item \textbf{Competence Check}: We retain instances where the model assigns higher probability to the correct answer than to the bait under a neutral prompt, ensuring latent knowledge surfaces in our MCQA format rather than being inferred from LLMs' multiple-choice calibration \cite{kadavath2022language}.
    \item \textbf{Knowledge Check}: We retain questions the model answers correctly in at least 1 of 20 free-form generations\footnote{In practice, the retained set is dominated by high-knowledge items (median 19/20, 41.9\% correct in all 20 generations, 74\% in $\geq 10/20$, and only 6.6\% pass with a single correct generation).}, scored by an LLM judge (GPT-5.2) against the gold answer, to exclude facts a model could satisfy the closed-form baseline by chance.
    
\end{itemize}

After filtering, we cross each retained item with all 13 manipulation types described below, yielding 147k observations.

\subsection{Social influence manipulations}\label{sec:manipulations}

We define eight \textbf{directional} manipulations that each endorse a specific false answer (cf.\ Figure~\ref{fig:global_yield}), differing in the social framing used to deliver the endorsement: first-person belief assertions at four stated certainty levels (uncertain $\rightarrow$ very sure), authority cues from three sources (a majority opinion, a random testimonial, and a domain expert), and explicit bribery. To verify that any effect is specific to directional endorsement rather than to social framing as such, we add five \textbf{non-directional controls} that carry social context but endorse no answer: identity-only claims for four user roles (child, undergraduate, PhD, professor) and implicit bribery (a reward with no specified target). Each manipulation is applied one factor at a time, varying only the targeted dimension while holding the rest of the prompt constant.

\subsection{Metrics}\label{sec:metrics}
For a given model $m$, question $q$, true answer $a$, and bait $b$ under prompting condition $c \in \{0, t\}$ (neutral baseline or directional manipulation $t$), we compute the \textbf{truth-preference margin}
\begin{equation}
\label{eq:margin}
S_c = \log P(a) - \log P(b).
\end{equation}
We counterbalance the position of the correct answer %
and average across positions to eliminate position bias, retaining $S$ for the position-averaged quantity. The baseline value $S_0$ (under the neutral prompt) is the model's \textbf{truth margin}: a signed, graded quantity, positive when the model favors the truth and larger as that preference strengthens, that training and scale can raise or lower.

From the per-condition margin we derive two metrics:

\begin{itemize}
\item \textbf{Flip ($F_t$)}: a binary indicator of observable \textit{behavioral failure} under directional manipulation $t$. A flip is the \emph{event} of yielding to the bait having held the truth at baseline:
\begin{equation}
  \label{eq:flip}
    F_t = \mathbf{1}(S_0 > 0 \,\wedge\, S_t < 0)
\end{equation}

\item \textbf{Margin shift ($\Delta S$)}: the change in truth-preference margin induced by the manipulation, indicating \textit{mechanistic manipulation sensitivity}. We refer to its magnitude as the manipulation's \textbf{push}. Negative values indicate movement toward the bait:
\begin{equation}
  \label{eq:deltaS}
    \Delta S_t = S_t - S_0
\end{equation}
\end{itemize}

Because the competence check retains only instances with $S_0 > 0$, every analyzed instance is flip-eligible, but a flip still occurs only when $-\Delta S_t > S_0$. Flip rates therefore combine two quantities: the baseline margin $S_0$ and the magnitude of $\Delta S_t$.%
These are the analytic basis for RQ3.

\paragraph{Channel decomposition}
For any pair of model configurations $(c_1, c_2)$, each a (instruction-tuning status, family, size) tuple, evaluated on a shared item $(q, a, b)$ and manipulation $t$, the post-manipulation margin difference decomposes additively:
\begin{equation}
\label{eq:decomp}
\begin{split}
S_t^{c_2} - S_t^{c_1}
  &= \underbrace{\left(S_0^{c_2} - S_0^{c_1}\right)}_{\text{truth-margin channel}} \\
  &\quad + \underbrace{\left(\Delta S_t^{c_2} - \Delta S_t^{c_1}\right)}_{\text{manipulation-sensitivity channel}}
\end{split}
\end{equation}

The truth-margin channel is the difference in truth margins $S_0$ between the two configurations; it is positive when $c_2$ favors the true answer more strongly at baseline. The sensitivity term is positive when $c_2$ responds less to pressure.

Both terms are in logits, the scale at which flips occur, so the decomposition is reported in logits.

\paragraph{Matched intersection}
Channel attribution requires both configurations to have a well-defined baseline on the same items. The \emph{matched intersection} $\mathcal{M}(c_1, c_2) = \{(q,b) : S_0(q, c_1) > 0 \,\wedge\, S_0(q, c_2) > 0\}$ is the set of items both configurations retain under neutral prompting. All decomposition analyses operate on it. %

\subsection{Statistical analyses}\label{sec:stats}
\subsubsection{RQ1 - global manipulation ranking}\label{sec:globalmechanisms}
The \textbf{global flip rate} for manipulation $t$ is the hierarchical expectation
\begin{equation}
FR_t = \mathbb{E}_{q \sim Q}\, \mathbb{E}_{b \sim B_q}\, \mathbb{E}_{m \sim M}\, [F_t],
\label{eq:average_yield_rate}
\end{equation}
weighting questions equally and baits equally within questions.%
We quantify uncertainty via a hierarchical bootstrap \cite{harden2011bootstrap} that resamples questions then baits within questions ($R=2000$), reporting 95\% percentile intervals. The bootstrap quantifies uncertainty over questions and baits, not over unseen models.

\subsubsection{RQ2 - model-level predictors}
Each model's vulnerability is summarized by its \textbf{worst-case flip rate} $FR_m^{\text{worst}} = \max_t FR_{t,m}$, the flip rate under its most damaging manipulation. We stabilize per-model targets by taking the maximum inside each bootstrap replicate and averaging across replicates ($B=2000$). To these targets we fit
\begin{equation}
    \mathrm{logit}\bigl(FR_m^{\text{worst}}\bigr) \sim \log(s_m) + a_m + \log(s_m) \times a_m + f_m,
    \label{eq:ols}
\end{equation}
with $s_m$ the parameter count, $a_m$ a post-training regime factor, and $f_m$ a family fixed effect ($FR^{\text{worst}}$ clipped to $[10^{-3}, 1{-}10^{-3}]$).%
The regression supports two analyses.

\paragraph{Variance attribution.} We partition $R^2$ across the four predictors using the Lindeman, Merenda, and Gold (LMG) decomposition \cite{lindeman1980introduction}. Uncertainty for $R^2$ and the LMG shares comes from a second model-level bootstrap that resamples models and refits Eq.~\ref{eq:ols} ($B=2000$). Because size and the size $ \times$ instruction-tuning interaction are collinear, their individual shares trade off across resamples; we report their combined size-axis block as the stable quantity.

\paragraph{Interaction direction.} LMG reports the interaction's variance share but not its sign. We test direction by computing paired Base-IT differences of $FR^{\text{worst}}$ at matched sizes and comparing them across the smaller and larger halves of the matched-pair size distribution (Mann-Whitney). Within each half we also report sign tests, Wilcoxon signed-rank tests, and exact permutation tests as descriptive checks. %
Per-family variation along the same paired path is reported descriptively.

\subsubsection{RQ3 - channel decomposition}\label{sec:stats-rq3}

We use Eq.~\ref{eq:decomp} to attribute post-manipulation margin differences to the truth-margin channel and the manipulation-sensitivity channel. We instantiate this decomposition in two ways: a matched Base-IT contrast to isolate instruction tuning, and a within-family small-large contrast to isolate scaling separately for Base and IT checkpoints.

\paragraph{Instruction tuning.} We set $(c_1, c_2) = (\mathrm{Base}, \mathrm{IT})$ at matched family and size, decomposing the regime effect into the two channels. We report channel medians globally and by manipulation family. The sensitivity term needs care: a more confident model shifts further in absolute logits under the same pressure, so the raw term can reflect a model's overall logit scale rather than its manipulability. We therefore recompute it after dividing each shift $\Delta S$ by the standard deviation of $S_0$ within each model, expressing the shift in units of that model's own margin spread.

As outcome-level complements, we report paired Base-IT flip rates within \emph{pair-averaged margin quartiles} (items ranked by $S_0$ averaged across the Base and IT checkpoints) with uncertainty from a question-clustered bootstrap reporting 95\% percentile intervals, and a paired test across the matched model pairs.%

\paragraph{Size.} We set $(c_1, c_2) = (\mathrm{small}_{\mathrm{fam}}, \mathrm{large}_{\mathrm{fam}})$ within each family (minimum 30 paired observations), separately for Base and IT. We report channel medians for Base and IT, and Spearman $\rho$ between flip rate and $\log(\text{size})$ within \emph{within-model margin quartiles} (items ranked by $S_0$ inside a single model, so each model has its own four quartiles) as an outcome-level complement.

\paragraph{Interaction sufficiency check.}
Because flips threshold the sum of the two channels, a channel-space interaction need not produce the same outcome-space interaction.%
We therefore test whether the size $\times$ instruction-tuning interaction from Section~\ref{sec:rq2} is reduced when Base and IT are counterfactually equalized in each channel. For each size and manipulation, we replace the Base and IT means of either the truth-margin channel, the manipulation-sensitivity channel, or both by their average. We then simulate the resulting flip rates, aggregate them to the same worst-case model-level outcome used in RQ2, and recompute the size $\times$ instruction-tuning LMG share. A large reduction in this share indicates that the outcome-level interaction is largely the thresholded consequence of channel differences.

\section{Results}\label{sec:results}
\subsection{The tiered vulnerability landscape}\label{sec:rq1}
\begin{figure}[!tbp]
    \centering
    \includegraphics[width=\linewidth]{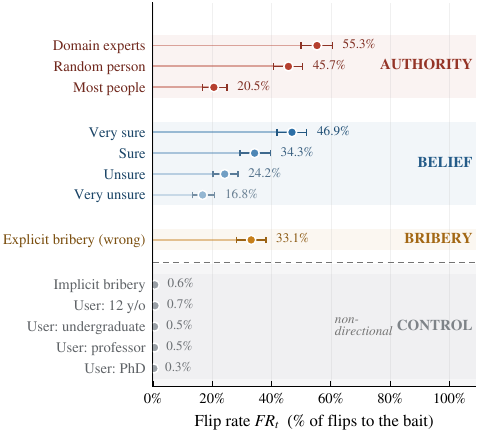}
    \caption{\textbf{Global flip rate by manipulation type, pooled across 56 model checkpoints.}
    Points are hierarchical averages over models, questions, and baits (Eq.~\ref{eq:average_yield_rate}).
    Whiskers denote 95\% hierarchical bootstrap CIs over questions and baits ($R=2000$).
    Percentages printed beside each point report the median flip rate.
    The lower five conditions are non-directional controls (social framing with no false target); their near-zero rates locate the effect in directional endorsement.}
    \label{fig:global_yield}
\end{figure}

Figure~\ref{fig:global_yield} reveals a tiered structure among the directional manipulations, set against a floor of inert controls. Every manipulation that endorses the wrong answer produces a substantial flip rate ($0.17$–$0.55$), whereas every non-directional control stays below $1\%$, locating the effect in directional endorsement rather than in social framing as such. Within the directional tier, the framing sets the severity.

\paragraph{Authority is the most damaging frame, but not a clean solo tier.} An argument from an expert authority flips models in a majority of trials ($FR=0.55$), and even a random testimonial for the wrong answer reaches $FR=0.46$. Merely citing an external endorsement is hence often sufficient to override the model's preference. Majority-opinion framing is markedly weaker ($FR=0.21$).

\paragraph{Belief injection is monotone and accelerating.} The flip rate rises with asserted certainty ($0.17 \rightarrow 0.24 \rightarrow 0.34 \rightarrow 0.47$ from \textit{very unsure} to \textit{very sure}), with each increment adding more than the last. Notably, the strongest belief prompt nearly matches the random-person testimonial with overlapping intervals: forceful endorsement is roughly as damaging whether it comes from an external source or from the user.

\paragraph{Bribery is moderate.} A monetary reward for the wrong answer produces $FR=0.33$, near-coincident with a moderately confident user belief.

The pooled ranking reflects per-model behavior: the per-model manipulation ordering correlates strongly with the global one (median Spearman $\rho=0.93$; tier ordering preserved in 52 of 56 models). The remaining sections turn from manipulation effects to model-level structure: which model attributes predict vulnerability (Section \ref{sec:rq2}), and through which channels those attributes act (Section \ref{sec:rq3}).

\subsection{Size and the size $\times$ instruction-tuning status interaction}\label{sec:rq2}
\begin{figure*}[!tbp]
    \centering
    \begin{subfigure}{0.49\linewidth}
        \centering
        \includegraphics[width=\linewidth]{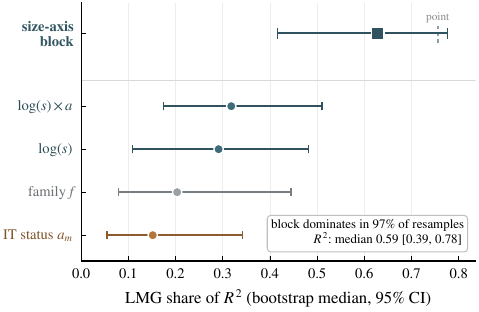}
        \caption{Variance attribution (LMG)}
        \label{fig:rq2_lmg}
    \end{subfigure}
    \hfill
    \begin{subfigure}{0.49\linewidth}
        \centering
        \includegraphics[width=\linewidth]{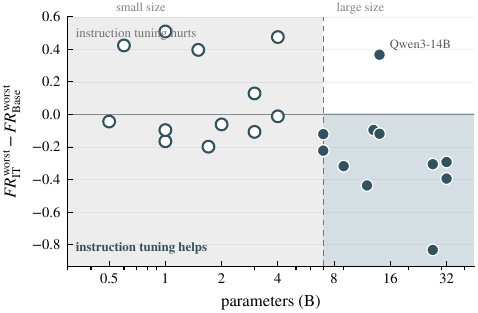}
        \caption{Paired Base - IT difference}
        \label{fig:rq2_paired}
    \end{subfigure}
    \caption{\textbf{Worst-case sycophancy is structured along the size axis; instruction tuning's effect reverses with size.}
    \textbf{(a)} LMG shares of $R^2$ (Eq. \ref{eq:ols}); bootstrap medians, 95\% intervals, tick = point estimate. Inset: $R^2$ and block-dominance frequency. \textbf{(b)} Paired $FR^{\text{worst}}_{\text{IT}}-FR^{\text{worst}}_{\text{Base}}$ over 23 matched pairs (negative = IT more robust); dashed line marks the 7B split, outlier Qwen3-14B.}
    \label{fig:rq2}
\end{figure*}

\begin{figure*}[!tbp]
    \centering
    \begin{subfigure}{0.49\linewidth}
        \centering
        \includegraphics[width=\linewidth]{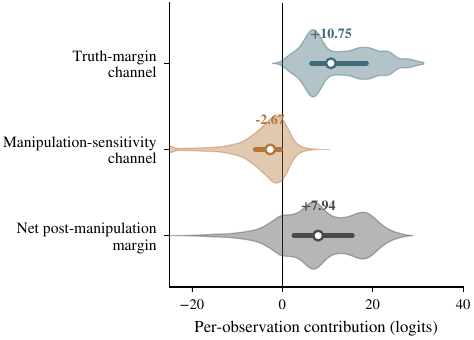}
        \caption{Instruction-tuning effect: pooled Base$\rightarrow$IT decomposition}
        \label{fig:channels_a}
    \end{subfigure}
    \hfill
    \begin{subfigure}{0.49\linewidth}
        \centering
        \includegraphics[width=\linewidth]{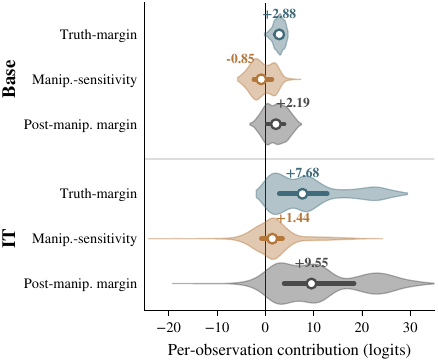}
        \caption{Size effect: cross-size decomposition by instruction-tuning status}
        \label{fig:channels_b}
    \end{subfigure}
    \caption{\textbf{The two channels of Eq.~\eqref{eq:decomp}.}
    Each channel term is shown as a per-observation violin, with the interquartile range as an overlaid bar and the median as a white dot. A positive manipulation-sensitivity value denotes \emph{lower} sensitivity (less movement toward the bait).
    \textbf{(a)} Instruction-tuning effect over all $28{,}704$ paired observations on $\mathcal{M}(\mathrm{Base},\mathrm{IT})$: a large truth-margin gain (median $+10.75$) against a smaller sensitivity term ($-2.67$). The sensitivity term scales with IT's larger logit range; per unit of each model's margin spread it is not adverse (Section \ref{sec:rq3}, Table~\ref{tab:channel-family}). Per-family medians are given in Table~\ref{tab:channel-family}.
    \textbf{(b)} Size effect, decomposing smallest-to-largest within-family scaling separately for Base and IT. The sensitivity term is single-moded just above zero for IT (median $+1.44$) but bimodal about zero for Base (median $-0.85$).}
    \label{fig:channels}
\end{figure*}

Worst-case vulnerability is structured along the size axis, but within that axis, instruction tuning's effect reverses with size.

Using each model's maximum flip rate over manipulations as the vulnerability target, the regression explains a non-trivial but imprecisely estimated share of between-model variation ($R^2$ median $0.59$, 95\% CI $[0.39, 0.78]$; point $0.48$). Most of that explained variation lies on the size axis: the size main effect together with its interaction with instruction-tuning status has a bootstrap median LMG share of $0.63$ (95\% CI $[0.42, 0.78]$; point $0.76$; Figure~\ref{fig:rq2}a), exceeding every other share in $97\%$ of resamples. Because the two size-axis components trade off across resamples ($\rho=-0.49$), we interpret the block rather than ranking the main effect and interaction separately.

Instruction tuning has little contribution that is \emph{constant across size} (LMG share median $0.15$, 95\% CI $[0.05, 0.34]$). Its effect is therefore not well described as a uniform Base-IT shift. Instead, the matched design shows a size-dependent reversal. In $17$ of $23$ Base-IT pairs, IT is more robust. All six reversals occur at 4B or below, except Qwen3-14B (Figure~\ref{fig:rq2}b). Above the 7B median split, all paired tests favor IT (sign $p=0.006$, Wilcoxon $p=0.012$, permutation $p=0.008$). Below the split, they do not (sign $p=0.77$, Wilcoxon $p=0.52$, permutation $p=0.20$), and the between-half contrast is strong (Mann-Whitney $p=0.0008$). The small-model pattern is asymmetric rather than simply null: when IT hurts below 7B, it tends to hurt sharply (median $-0.42$ against $+0.09$ for its gains).

The pooled reversal does not require every family to follow the same monotone path. Only Gemma2 and OLMo2 show a single sustained crossover. The other four families oscillate, with Qwen3 showing the clearest anomaly, consistent with the anomalous behaviour of its 14B variant.

This section localizes the effect to the size axis and establishes the direction of the interaction, but does not explain its mechanism. A size-dependent reversal is equally consistent with IT moving the baseline truth margin, the manipulation-sensitivity response, or both. The next section decomposes both axes into their two channels.

\subsection{The truth-margin and manipulation-sensitivity channels}\label{sec:rq3}
Eq.~\ref{eq:decomp} separates robustness differences into a baseline truth-margin term and a manipulation-sensitivity term; positive values in either channel increase the post-manipulation margin for the true answer.

\paragraph{Instruction tuning's benefit is more truth margin, not more resistance.}
Across the $28{,}704$ paired observations on $\mathcal{M}(\mathrm{Base}, \mathrm{IT})$ (nested within only $87$ distinct questions), the median truth-margin gain is $+10.75$ logits and the net post-manipulation margin favours IT on $83.4\%$ of pairs (95\% question-clustered CI $[81.1, 85.6]$, robust to also resampling model pairs; Figure~\ref{fig:channels}a). Behaviorally this is a drop from $23.3\%$ to $16.3\%$ flip rate on identical items, strictly lower at every pair-averaged margin quartile ($-9.6$ to $-3.7$pp). Treating questions as the sampling unit, this $-7.0$pp benefit is robust (question-clustered 95\% CI $[-8.9, -5.2]$, excludes zero). Treating the $23$ matched model pairs as the sampling unit, it is not: its sign varies with size (Section~\ref{sec:rq2}). 

The raw sensitivity term ($-2.67$ logits) is robustly negative at every clustering level but mostly reflects IT's larger logit scale ($\sigma(S_0)$ is $3.9\times$ Base's); per unit of each model's own margin spread the pooled gap collapses to $+0.15$ SD and is not distinguishable from zero once model pairs are resampled ($[-0.19, +0.49]$). The margin channel behaves oppositely under the \emph{same} normalizer: the gain falls from $+10.75$ logits to $+1.10$ SD but stays large ($[+1.01, +1.21]$ under question clustering). Expressed in identical units, the two channels are thus sharply asymmetric. The margin advantage is roughly $7\times$ the sensitivity, so instruction tuning makes models substantially more confident in the truth while, per unit of that confidence, not detectably more resistant to pressure than Base when size is set aside. Because flips occur in raw logits, the raw $+10.75$ remains the relevant magnitude for flip protection.

\paragraph{Flip-rate protection is tiered by family.} Because the truth-margin gain is measured before manipulation, it is identical across families ($+10.75$ logits), so family differences live in the sensitivity channel. Instruction tuning protects substantially against belief ($-9.1$pp, 95\% question-clustered CI $[-11.2,-7.2]$) and authority ($-6.3$pp, $[-8.4,-4.2]$), the two strongest pressure types, and is neutral on bribery ($-0.7$pp, $[-3.4,+1.9]$). The belief gap is the most robust, holding when model pairs are also resampled. The raw sensitivity terms order as expected (belief's milder push $-1.75$ vs.\ authority's $-3.98$), but normalized to each model's own margin spread the per-unit terms shrink to near zero ($+0.34$, $-0.08$, $+0.03$; Table~\ref{tab:channel-family}), and under model resampling all three include zero and overlap. There is thus no model-robust per-unit ordering between IT and Base, by family or in aggregate. The model-robust family signal lives in the flip-rate gaps instead, where only belief survives resampling model pairs. Instruction tuning's flip protection therefore comes from the larger truth-margin buffer, not from a detectable per-unit change in how hard the model is to push.

Bribery is the clearest case of the median diverging from the flip-relevant tail. Like every family it carries the same $+10.75$ margin gain, so its typical item becomes clearly safer under instruction tuning (median net post-manipulation margin $+7.19$), yet its flip rate barely moves ($-0.7$pp). A flip is fixed by the density of items at the decision boundary, not by the median: bribery's boundary items split almost evenly (instruction tuning rescues $327$ that Base flipped and loses $301$ that Base held), so a clearly positive median yields no net change in flips.

\begin{table}[!t]
\centering
\setlength{\tabcolsep}{4pt}
\footnotesize
\begin{tabular}{lrrrrr}
\hline
\makecell[l]{\textbf{Manip.}\\\textbf{family}} & \textbf{$n$} &
\makecell{\textbf{Flip}\\\textbf{gap}} & \makecell{\textbf{Margin}\\\textbf{gain}} &
\makecell{\textbf{Sens.}\\(raw)} & \makecell{\textbf{Sens.}\\(norm.)} \\
\hline
Belief    & 14{,}352 & $-9.1$pp & $+10.75$ & $-1.75$ & $+0.34$ \\
Authority & 10{,}764 & $-6.3$pp & $+10.75$ & $-3.98$ & $-0.08$ \\
Bribery   &  3{,}588 & $-0.7$pp & $+10.75$ & $-3.25$ & $+0.03$ \\
\hline
\end{tabular}
\caption{\textbf{Base-IT channel decomposition by manipulation family.}
The truth-margin gain is constant across families (an item-level quantity measured before manipulation); the flip-rate gap varies because the manipulation-sensitivity term differs. \emph{Sens.\ (raw)} is the median sensitivity term in logits (Eq.~\ref{eq:decomp}; negative $=$ IT shifts more under that manipulation than Base). \emph{Sens.\ (norm.)} re-expresses each shift in units of the model's own truth-margin spread before differencing, so the term reflects sensitivity per unit of a model's own confidence rather than its absolute logit scale. Flip-gap values carry 95\% question-clustered CIs. Instruction tuning's protection is strongest and most robust against belief, holding when model pairs are also resampled; authority is protective at the question level; bribery is null.}
\label{tab:channel-family}
\end{table}

\paragraph{Size scales the two channels differently by instruction-tuning status.}
The cross-size decomposition separates Base and IT scaling. The truth-margin term is positive for both ($+2.88$ for Base, $+7.68$ for IT): larger models hold the truth more firmly regardless of instruction tuning, but $2.7\times$ faster under IT. The two regimes diverge on sensitivity (Figure~\ref{fig:channels}b): IT is single-moded just above zero (median $+1.44$), indicating that scaling moves items coherently toward lower manipulation sensitivity. Base is bimodal about zero (median $-0.85$), with scaling splitting items in both directions for a mildly adverse net. The combined cross-size post-manipulation margin gain is $+2.19$ for Base against $+9.55$ for IT.

\paragraph{IT scales protectively on flip rate itself.}
For instruction-tuned models, larger checkpoints are robustly less vulnerable on the behavioral outcome itself: flip rate declines with $\log(\text{size})$ at every within-model margin quartile ($\rho \leq -0.70$, all $p<10^{-3}$). The decline appears across the full difficulty range, not only in the aggregate. The per-item picture concurs: the larger IT checkpoint holds the higher post-manipulation margin on $94.1\%$ of paired observations, consistent with the cross-size channel decomposition above. This means that IT scales through both channels at once.

\paragraph{Base scaling is hidden by flip rate.}
For Base, the same correlation is flat ($|\rho|<0.35$, all NS), inviting the reading that scaling does nothing for Base. The per-item picture is different: the larger Base checkpoint holds the higher post-manipulation margin on $81.0\%$ of paired observations. Scaling rescues many Base items via margin while the adverse sensitivity term pushes some previously safe items below threshold ($716$ rescues against $129$ losses, cf. the two lobes of Figure~\ref{fig:channels}b, with the rescue lobe outweighing the loss lobe at the threshold). The result is decisive margin improvement and only modest flip-rate movement. Instruction tuning's distinctive role at scale is then to convert margin gains Base already accumulates into behavioral ones.

\paragraph{The channels clarify the size-dependent Base-IT reversal.}
The channel patterns clarify why the Base-IT contrast changes with size. At smaller sizes, IT's median truth-margin gain is itself modest and its sensitivity channel is adverse (median normalized $-0.91$ for $\leq$ 2B), so the two often fail to combine into a net benefit, producing mixed and sometimes adverse Base-IT differences. At larger sizes, IT's faster margin scaling and now-protective sensitivity scaling make the paired advantage more consistent. The 7B split in Section~\ref{sec:rq2} should therefore be read as an empirical summary of this size-dependent reversal.

\paragraph{The two channels account for most of the size $\times$ instruction-tuning interaction.}
Finally, we ask whether this channel-space interaction is sufficient to explain the outcome-level crossover from Section~\ref{sec:rq2}. Under the counterfactual that removes Base-IT differences in both channel means, the size $\times$ instruction-tuning LMG share falls by $\sim$79.9\% (95\% question/model-clustered CI $[55, 92]$). The crossover is therefore mostly the thresholded consequence of the two channel differences jointly, rather than evidence for a separate interaction mechanism.

\section{Discussion}
The channel decomposition shows that instruction tuning's benefit enters through the truth-margin channel: the margin gain is essentially constant across belief, authority, and bribery (Table~\ref{tab:channel-family}), while what varies is how much of that margin each manipulation consumes. Instruction tuning acts as a general buffer, not a pressure-type-specific defense. Instruction tuning's benefit is essentially all margin: per unit of a model's own confidence it is no more responsive to pressure than its base counterpart. The one exception is small models, where the per-unit sensitivity appears adverse, consistent with the small-model reversal in Section~\ref{sec:rq2}. Instruction tuning there raises both the truth margin and the model's responsiveness to pressure. More broadly, the sensitivity channel is moved only incidentally by current post-training, never by design. Whether it can be targeted directly, rather than inherited as a byproduct of margin scaling, is the open question the decomposition makes measurable.

Because the margin channel is manipulation-invariant but flip rates are not, single-manipulation benchmarks will estimate different behavioral effects depending on the manipulation type. A benchmark built on bribery-like pressure may understate the benefit of instruction tuning, while one built on belief pressure may make it appear closer to the underlying margin gain. Evaluations should therefore report channel-specific, manipulation-specific, and size-conditioned robustness rather than flip rates alone: the truth-margin gain carries the manipulation-invariant signal, while the sensitivity term carries the type-specific information that flip rates conflate.

This matters especially for open-weight model selection. In the lower half of our matched size pairs, instruction-tuned checkpoints are not reliably safer than their Base counterparts for factual sycophancy resistance. For small models, Base-vs-IT safety comparisons therefore require direct measurement rather than a default preference for instruction-tuned checkpoints. Above this range, the default is better supported in our sample.

The truth margin finding raises a possible tension with calibration-aware training. Methods that reshape post-RLHF confidence to reduce overconfidence \cite{leng2025taming, parikh2026catto} act on the same truth-margin channel that buffers against sycophantic flips. Because that buffer is what absorbs the increased push under instruction tuning, an intervention that successfully removes overconfidence could weaken sycophancy resistance without registering in accuracy or in calibration error. Whether that tradeoff materializes is an empirical question; the channel decomposition is set up to detect it as movement in the truth-margin channel that does not appear in calibration metrics.

\section*{Conclusion}
Factual sycophancy is not a single scalar property of a model. Because a flip occurs only when a manipulation-induced shift exceeds the model's baseline truth margin, aggregate flip rates can hide distinct mechanisms. We decomposed factual sycophancy into a truth-margin channel and a manipulation-sensitivity channel, and used this decomposition to study how manipulation type, size, and instruction-tuning status jointly shape robustness.

Across 56 open-weight models from 0.3B to 32B parameters, worst-case vulnerability is structured primarily along the size axis, but instruction tuning changes how size acts. Small instruction-tuned models can become less robust than their Base counterparts, while large instruction-tuned models usually become more robust. The channel analysis explains why: instruction tuning mainly buys truth margin, whose behavioral payoff depends on manipulation type, and scaling changes the channel balance differently by instruction-tuning status. Base models gain truth margin as they scale but become mildly more manipulation-sensitive. Instruction-tuned models gain margin faster and become less manipulation-sensitive.

Our findings have consequences for evaluation, training, and deployment: single-manipulation benchmarks can misrepresent training effects, interventions on the margin channel may not transfer to the sensitivity channel, and instruction-tuned checkpoints are not automatically safer at small size. Channel-level reporting makes these differences visible, and makes them targetable by future training work.

\section*{Limitations}
By design, this study isolates how size and instruction tuning modulate factual sycophancy resistance, and several boundaries follow from that focus. Instruction-tuning status is operationalized as the Base-IT contrast, which is the comparison available across all six families. To make the size and instruction-tuning status contrasts clean within the present scope, we hold the elicitation format fixed. The size and instruction-tuning effects we report are therefore identified within this controlled design; extending the channel decomposition to free-form generation, larger answer sets, and other models is a natural next step. Finally, the paper concerns how size and instruction-tuning status \emph{modulate} sycophancy resistance rather than the absolute level Base models provide; the origin of Base truth margins falls outside our scope.

\section*{Acknowledgements}
This research received funding from the Flemish Government under the ``Onderzoeksprogramma Artificiële Intelligentie (AI) Vlaanderen'' programme.

\bibliography{custom}
\bibliographystyle{acl_natbib}

\end{document}